%% file: root.tex
\documentclass[letterpaper, 10 pt, conference]{ieeeconf}  

\IEEEoverridecommandlockouts                              

\usepackage{graphics} 
\usepackage{epsfig} 
\usepackage{mathptmx} 
\usepackage{times} 
\usepackage{amsmath} 
\usepackage{amssymb}  
\usepackage{booktabs}
\usepackage{balance}
\usepackage{threeparttable} 
\usepackage{graphicx}
\usepackage{tikz}                              
\usetikzlibrary{arrows.meta,calc,positioning}  
\usepackage{algorithm}
\usepackage{algpseudocode}
\usepackage{cite}
\usepackage[table]{xcolor}
\definecolor{oursrow}{HTML}{CCFFCC}
\definecolor{projlink}{HTML}{4338CA}   
\newcommand{\projsize}{\normalsize}  
\newcommand{\projurl}{https://advanced-robotic-manipulation.github.io/websites/hapticwam/}
\usepackage{hyperref}
\hypersetup{hidelinks}
\usepackage{censor}
\usepackage{comment}

\title{\LARGE \bf
HapticWAM: Distilling Imagined Touch into a World--Action Model without Inference-Time Tactile Sensing
}

    \author{
    \parbox{16.5cm}{\centering
        Mikhail Sannikov*, Ilya Mikhalchuk*, Konstantin Gubernatorov*, Petr Kovalev, \\
        Ogunwoye Faith Oluwatobi, and Dzmitry Tsetserukou\\[5pt]
        {\projsize\bfseries Project page:\enspace
         \href{\projurl}{\textcolor{projlink}{\expandafter\nolinkurl\expandafter{\projurl}}}}
    }
    \thanks{*Denotes equal contribution.}
    \thanks{All authors are with the Intelligent Space Robotics Laboratory, Skolkovo Institute of Science and Technology, Moscow, Russia. \tt \{{Mikhail.Sannikov}, {Ilia.Mikhalchuk}, {Konstantin.Gubernatorov}, {Petr.Kovalev}, {Faith.Ogunwoye},{D.Tsetserukou}\} @skoltech.ru}
}

\begin{document}

\setlength{\textfloatsep}{7pt plus 2pt minus 2pt}
\setlength{\dbltextfloatsep}{7pt plus 2pt minus 2pt}
\setlength{\floatsep}{7pt plus 2pt minus 2pt}
\setlength{\dblfloatsep}{7pt plus 2pt minus 2pt}
\setlength{\intextsep}{7pt plus 2pt minus 2pt}
\setlength{\abovecaptionskip}{3pt}
\setlength{\belowcaptionskip}{0pt}

\maketitle
\thispagestyle{empty}
\pagestyle{empty}

\input{Abstract}

\section{Introduction}
\input{Introduction}

\section{Related Work}
\input{Related_work}

\section{Method}
\input{Method}

\section{Experiments}
\input{Experiments}


\section{Conclusion}
\input{Conclusion}

\balance
\bibliographystyle{IEEEtran}
\bibliography{references}

\end{document}

%% file: Abstract.tex
\begin{abstract}
Contact-rich manipulation requires estimating forces, slip and contact
geometry that can remain ambiguous in scene images. Optical tactile sensors
provide both visual observations of the contact surface and mechanical
measurements, yet learning from these signals raises two challenges:
representing contact beyond appearance and transferring its benefits to a
policy that does not require fingertip observations at deployment.
We introduce \textbf{HapticWAM}, a world--action model that combines
heterogeneous tactile encoding, structured contact prediction and
teacher--student distillation. Its teacher encodes gel images together with
deformation, shear, distributed forces, resultant wrench and derived contact
state into a frozen video backbone. Rather than predicting tactile pixels
alone, the model jointly generates actions and a contact package describing
future events and mechanics. Anticipatory Contact Coupling uses the previously
imagined package to condition attention, preserving a contact-related input
when direct tactile observations are unavailable. Haptic-Imagination
Distillation transfers both contact futures and action predictions to a
student that retains the generative contact head but removes its fingertip
input branches. On a real-world setup, across three contact-rich pick-and-place tasks, HapticWAM Student achieves
a 77\% per-task mean success rate (41 of 50 starts, 82\% pooled), reaching 95\% on one of the tasks, outperforming the evaluated teacher and baseline
configurations.
\end{abstract}

%% file: Introduction.tex
\begin{figure}[!t]
\centering
\includegraphics[width=\columnwidth]{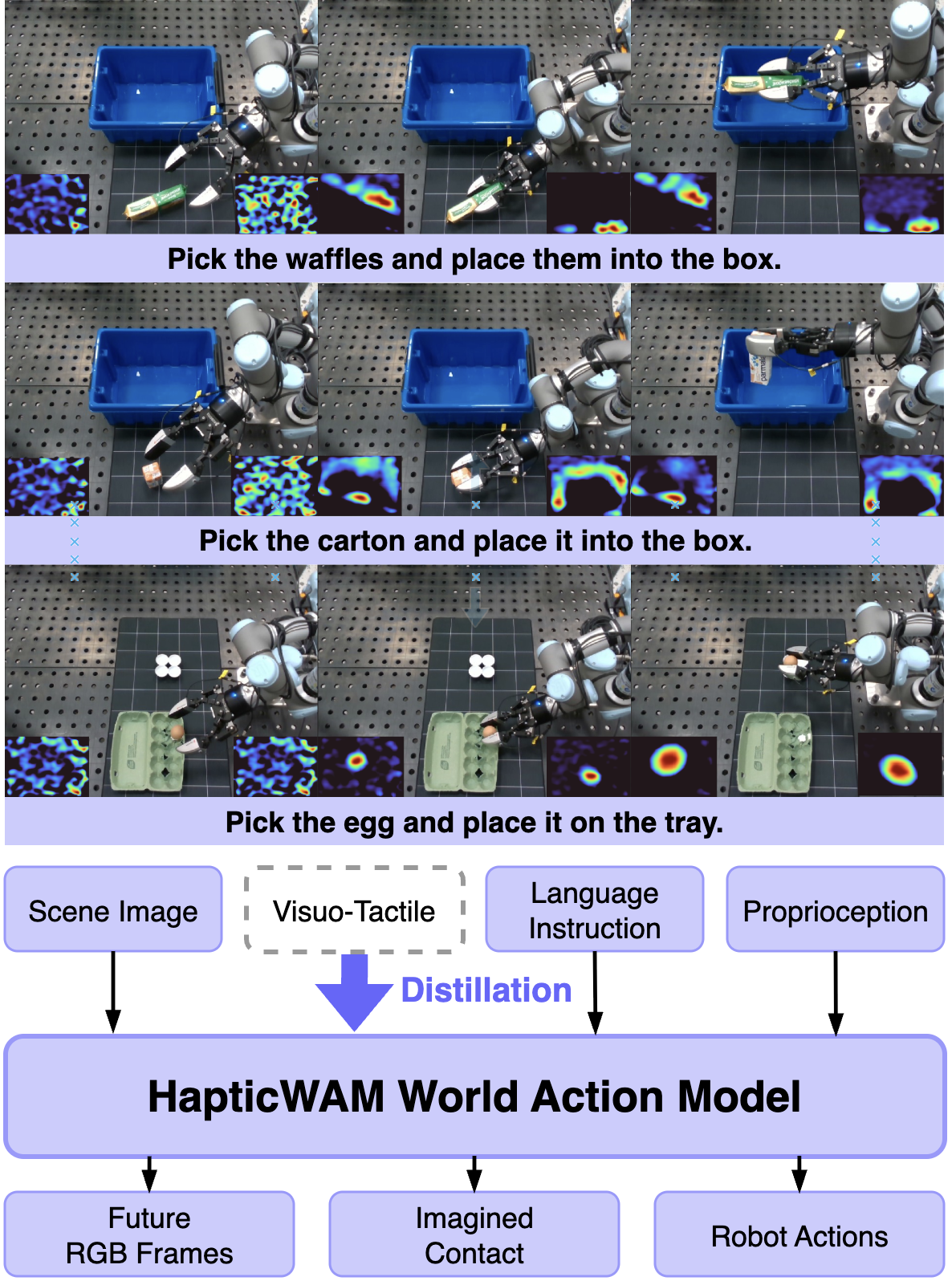}
\caption{HapticWAM encodes the full range of contact modalities an optical-tactile sensor reports, rather than gel appearance alone. Our approach leverages Haptic Imagined Distillation to preserve inference-time tactile-aware reasoning while eliminating optical-tactile sensors as input through distillation.}
\label{fig:teaser}
\end{figure}

Picking up a fragile object requires more than reaching the correct pose.
The fingers must establish contact, maintain enough force to prevent slip
and release the object at the target. Scene images reveal object geometry
and motion but may not distinguish a stable grasp from light contact,
particularly when the gripper occludes the interaction. Optical tactile
sensing complements vision with local contact appearance and mechanics.
The central question is how to learn from this richer information while
reducing the deployed policy's dependence on direct fingertip measurements.

Vision--language--action models
(VLAs)~\cite{brohan2023rt2,kim2024openvla,black2024pi0,intelligence2025pi05visionlanguageactionmodelopenworld}
map observations and task instructions to action chunks. Tactile VLAs add
contact observations to this interface
\cite{huang2025tactile,zhang2026vtla,bi2026vla,team2026n_0_vtla,gubernatorov2026hapticvla}.
These policies can learn anticipatory behavior without an explicit world
model. However, action supervision alone does not require them to expose a
prediction of the contact events or mechanical changes associated with a
candidate motion. World--action models (WAMs) jointly generate actions and future
observations
\cite{nvidia2025cosmos,wam_zeroshot,wvam,tang2026predictive,fastwam},
providing a framework in which these quantities can be learned explicitly.
Predicting visual futures by itself does not directly supervise grip force,
contact geometry or slip.

For optical tactile sensors, the available information also extends beyond
an image of the gel. Native outputs include deformation, indentation, shear,
distributed forces and resultant wrench; contact masks, centres of pressure
and slip indicators can be derived from these streams. An appearance-only
input leaves this mechanical structure implicit in the image representation.
Conversely, reconstructing future tactile pixels allocates prediction
capacity to appearance as well as task-relevant contact changes.
Prior tactile world models already explore image-based and structured
predictions~\cite{dreamtac,tactile_wam,team2026n_0,xue2026hitac}.
We build on this progression by connecting heterogeneous native sensor
inputs to an explicit contact-future representation that survives removal
of tactile observations from the learned policy.

This removal introduces a second problem. A policy trained to condition on
measured contact cannot simply retain the same observation interface when
those measurements are unavailable. Privileged-modality distillation offers
a way to transfer supervision from a tactile teacher to a student with fewer
inputs. Here the student must predict contact from visual approach geometry,
robot state and an arm-side wrench estimate. It cannot recover arbitrary
unobserved contact conditions, but it can be trained to represent the
teacher's predicted contact consequences alongside its actions. Matching
these futures provides an explicit supervision target beyond copying
actions; whether it improves training over action-only distillation is a
separate empirical question.

We propose \textbf{HapticWAM} to connect these representation and transfer
problems. A Heterogeneous Haptic Tokenizer (HHT) maps tactile appearance and
mechanics into a frozen video diffusion transformer. The model jointly
generates an action chunk and a structured package of contact events,
mechanical changes and contact geometry. Anticipatory Contact Coupling
(ACC) conditions attention on the package imagined at the previous replan,
so its contact-related input need not come from a live tactile stream.
Haptic-Imagination Distillation (HID) then matches the teacher's contact
futures and action predictions in a student that removes its tactile
observation frames but retains contact generation.
Figure~\ref{fig:teaser} illustrates the evaluated tasks.
The deployment claim concerns tactile \emph{policy inputs}: the robot
comparisons retain mounted sensors and the shared controller's tactile guards.

Our contributions are:
\begin{enumerate}
\item \textbf{Heterogeneous Haptic Tokenization} combines gel appearance, mechanical fields and derived contact state (the full modality range available from DM-Tac W2L sensors) in a common latent-frame interface, supporting joint structured-contact and action prediction.
\item \textbf{Anticipatory Contact Coupling} uses current leading signals
and the previous imagined contact package to bias attention toward haptic
tokens, including when fingertip observations are unavailable.
\item \textbf{Haptic-Imagination Distillation} combines teacher
initialization, future-contact and event matching, action matching and
ground-truth supervision in a student without tactile inputs.
\item \textbf{A visuo-tactile dataset and real-wrold robot evaluation} support comparisons
of the teacher, student and visual baselines on shared pick-and-place
starts, together with a deployment-time contact-generation intervention.
\end{enumerate}

%% file: Related_work.tex
\subsection{World--Action Models}
Imitation policies~\cite{chi2023diffusion} and VLAs
\cite{kim2024openvla,black2024pi0} predict actions from observations and task
context. World--action models add an explicit generative future. Video
foundation models~\cite{nvidia2025cosmos} support zero-shot robot
control~\cite{wam_zeroshot}, value-guided generation~\cite{wvam} and joint
predictive reasoning and action learning~\cite{tang2026predictive}.
Efficient inference is an active direction~\cite{fastwam}, and
Being-H0.7 goes further by replacing frame generation with learnable
latent queries between perception and action, arguing that pixel-space
prediction is a costly and indirect substrate for control~\cite{being_h07}.
Whether the future is rendered as pixels or compressed into latents, it is
learned from video: a visually plausible future does not by itself specify
the forces and contact transitions needed to execute it. This distinction
matters when different contact states produce similar scene images.

HapticWAM shares the view that pixel fidelity is not the objective, but
rather than removing the explicit future it changes what that future
describes: the pretrained visual backbone is retained while the prediction
space is extended with contact events, force and displacement changes, and
contact geometry. Video provides scene context, whereas contact streams
provide direct targets for the interaction quantities represented by the
model. The distinction is a choice of supervision and representation, not
a claim that action-only policies cannot learn anticipatory behavior or
that explicit contact generation alone guarantees better control.

\subsection{Tactile World Models}
Dream-Tac combines tactile prediction with contact-aware
attention~\cite{dreamtac}, while Tactile-WAM uses asymmetric attention over
tactile tokens~\cite{tactile_wam}. VT-WAM and TacForeSight couple
visual--tactile prediction with force guidance~\cite{vt_wam,tacforesight};
TacPAC corrects actions using discrepancies between expected and observed
touch~\cite{ma2026tacpac}. Other approaches predict tactile futures for
planning~\cite{flowtouch,vt_world_models}, imagine touch in a
VLA~\cite{dreamtacvla}, or model wrist forces~\cite{he2026fawam}.
Structured prediction is also established: N$_0$-TWAM uses a force-based
representation~\cite{team2026n_0}, and HiTac-WAM factorizes contact,
deformation and slip~\cite{xue2026hitac}. These precedents motivate treating
touch as more than a stream of images.

Our focus is the connection between heterogeneous tactile measurements,
future-contact generation and deployment without direct tactile policy
inputs. HHT gives appearance, mechanical fields and contact state distinct
encoding paths rather than requiring a single image encoder to infer all
of them. The output package represents contact events and mechanical changes
instead of gel appearance alone. ACC then uses the model's previously
predicted package, making this attention-conditioning input available to
the student after its fingertip observation branches are removed.
The proposed contribution is this combination, not the first use of force
prediction, structured touch or contact-aware attention.

\subsection{Privileged-Modality Distillation}
Privileged teachers transfer information unavailable to deployed students.
In manipulation, PTLD transfers tactile representations for sim-to-real
control~\cite{chen2026ptld}, FD-VLA distills force supervision~\cite{fdvla},
and TacImag predicts tactile observations from vision and
proprioception~\cite{zhang2026imagining}. HapticVLA~\cite{gubernatorov2026hapticvla} distills a
tactile-conditioned reactive policy into a student that predicts a tactile
token. These methods address observation
asymmetry, but differ in what the student retains: a teacher representation,
an imitated action or a predicted sensory quantity.

HID retains a generative \emph{contact-future} head and supervises its
mechanics and event predictions alongside the action field. Event saliency
and predicted teacher uncertainty weight contact matching, while
ground-truth supervision anchors the student to demonstrations.
Recorded teacher and earlier-student rollouts broaden the training
distribution, following the dataset-aggregation motivation
\cite{ross2011dagger}. Fully on-policy distillation, in which the teacher
supervises samples the student itself generates, outperforms off-policy
matching in language-model settings and, with reward extrapolation, lets a
student exceed its teacher~\cite{beyond_teacher}; HID approximates this
setting offline from recorded rollouts rather than collecting new on-policy
data for each student, and does not extrapolate beyond the teacher. This gives the student an explicit predicted-contact interface
without requiring fingertip observations. The present robot ablation
tests use of that interface at deployment, rather than isolating the
training benefit of contact targets over action-only matching.

%% file: Method.tex
\subsection{Problem Formulation}
\label{sec:method-overview}
At each replan $t$, the policy receives a scene image $I_t$, robot state
$q_t$, task text and an arm-side wrench history $w_{t-k:t}$ over the last
$k$ control ticks. The wrist signal is
estimated from motor currents, not measured by a wrist force sensor.
The teacher additionally observes gel images $G_t^f$, mechanical fields
$X_t^f$ and derived contact states $\kappa_t^f$ from two fingertips indexed
by $f\in\{1,2\}$; superscripts $\mathrm T$ and $\mathrm S$ denote teacher
and student throughout:
\begin{align}
o_t^{\mathrm{T}} &= (I_t,q_t,w_{t-k:t},\{G_t^f,X_t^f,\kappa_t^f\}_f),\label{eq:obs}\\
o_t^{\mathrm{S}} &= (I_t,q_t,w_{t-k:t}).\label{eq:obs-student}
\end{align}
Task text is supplied as shared conditioning in addition to these sensor inputs.
Both policies predict a chunk of end-effector pose deltas and absolute
gripper apertures. The student removes tactile observations, while retaining
vision, proprioception and the arm-side wrench estimate.

\begin{figure*}[t]
    \centering
    \includegraphics[width=\textwidth]{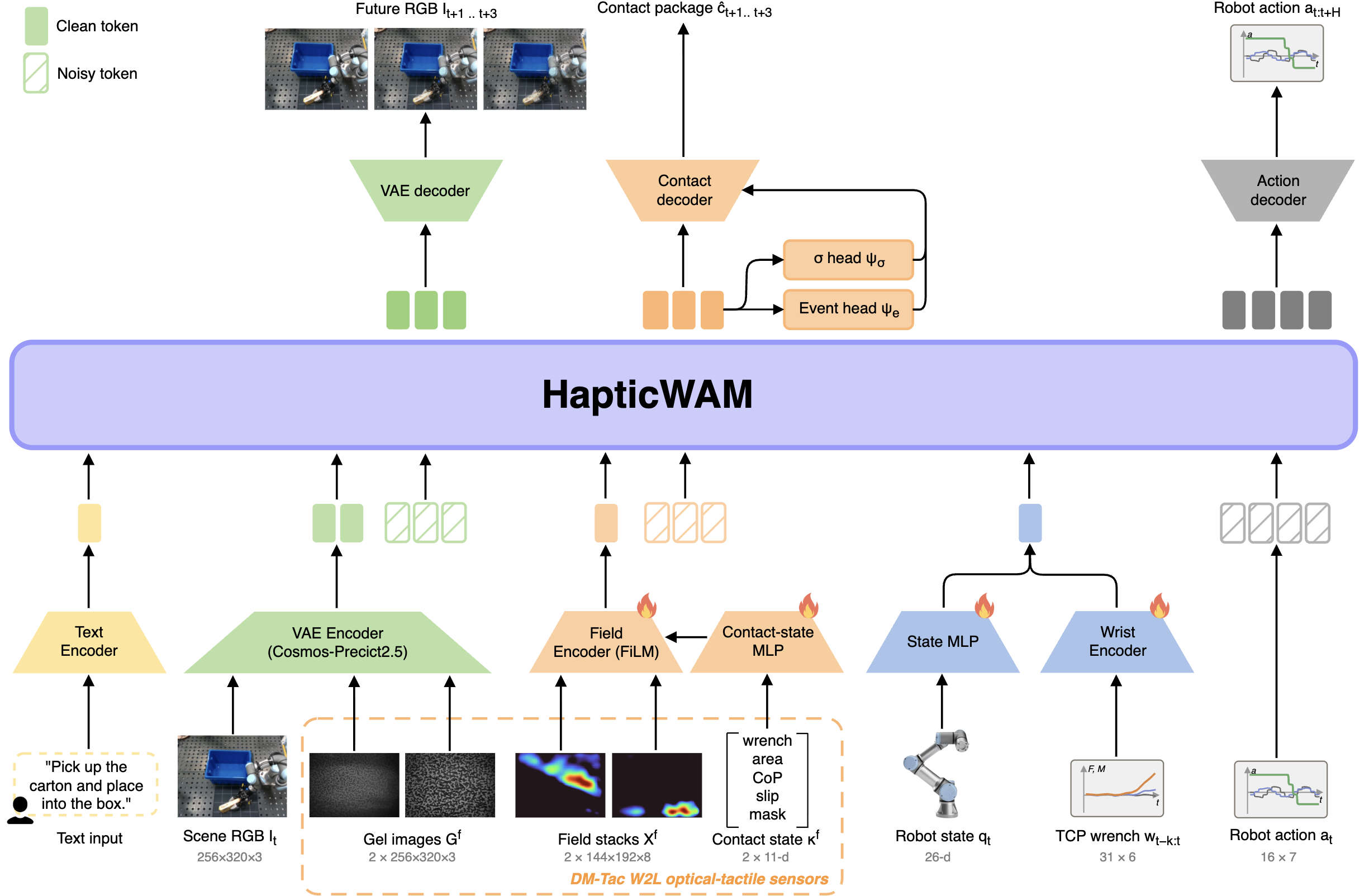}
    \caption{HapticWAM framework overview. \emph{Inputs:} the scene image $I_t$, robot state $q_t$ and arm-side wrench history $w_{t-k:t}$, with task text and the previously executed action chunk as conditioning; the teacher additionally observes both fingertips' gel images $G^f_t$, mechanical fields $X^f_t$ and contact states $\kappa^f_t$. Images pass through the frozen video VAE, mechanical fields through a convolutional encoder, and robot state and wrench history through MLP and causal-convolution encoders. \emph{Outputs:} future video, a structured contact package per future step, and a chunk of end-effector pose deltas and gripper apertures. The student is the same model without visuo-tactile input modalities from optical-tactile sensors.}
    \label{fig:arch}
\end{figure*}

A frozen Cosmos video diffusion transformer~\cite{nvidia2025cosmos} processes
observation, video, contact and action groups in one latent sequence. Its
backbone is a diffusion transformer (DiT)~\cite{peebles2023scalable}: a stack of
28 identical blocks that denoise spatio-temporal latent tokens with
self-attention, cross-attention to the task-text embedding and a
feed-forward multilayer perceptron (MLP). The sequence is
\begin{equation}
z=[z_{\mathrm{obs}},z_{\mathrm{video}},z_{\mathrm{contact}},z_{\mathrm{action}}].
\label{eq:layout}
\end{equation}
Observation frames are fixed conditioning. Generated groups are jointly
denoised using rectified flow: a clean latent $x_0$ is mixed with Gaussian
noise $\epsilon$ at noise level $\tau\in[0,1]$, the network $v_\theta$
regresses the velocity target $v^\star$, and $\hat x_0$ is the denoised
estimate:
\begin{align}
x_\tau&=(1-\tau)x_0+\tau\epsilon,\label{eq:flow}\\
v^\star&=\epsilon-x_0,\label{eq:flow-target}\\
\hat x_0&=x_\tau-\tau v_\theta(x_\tau,\tau,o_t).\label{eq:flow-prediction}
\end{align}
Only low-rank adaptation (LoRA) adapters~\cite{hu2021lora}, observation encoders, ACC and the contact readout heads
are trainable. The previously executed action chunk also conditions the
backbone. Figure~\ref{fig:arch} summarizes the architecture.

\subsection{Heterogeneous Haptic Tokenization}
\label{sec:method-hht}
HHT maps heterogeneous observations into the backbone's latent-frame
interface. Each fingertip supplies gel appearance and eight mechanical
channels: two deformation components, depth, two shear components and three
distributed-force components. Its 11-dimensional contact state contains
six resultant-wrench components, area, two centre-of-pressure coordinates,
slip and mask fraction. The frozen variational autoencoder (VAE)
encodes the gel images; a learned pointwise convolution fuses the two
fingertips into one appearance frame. A convolutional encoder
$\mathcal E_X$ processes mechanical fields with contact-state FiLM
modulation~\cite{perez2018film}, whose scale $\gamma$ and shift $\beta$ an MLP computes from $\kappa_t^f$:
\begin{equation}
z_t^f=(1+\gamma(\kappa_t^f))\odot\mathcal E_X(X_t^f)+\beta(\kappa_t^f),
\label{eq:hht-mech}
\end{equation}
where $\odot$ is the elementwise product. The modulation MLP is
zero-initialized. A causal temporal convolution~\cite{lea2016temporal}
$\mathcal E_w$ encodes the wrench history, and an MLP encodes robot state. The student
deletes the fused gel and mechanics observation frames, reducing the
sequence from 14 to 12 latent frames.

\begin{figure*}[t]
    \centering
    \includegraphics[width=0.9\textwidth]{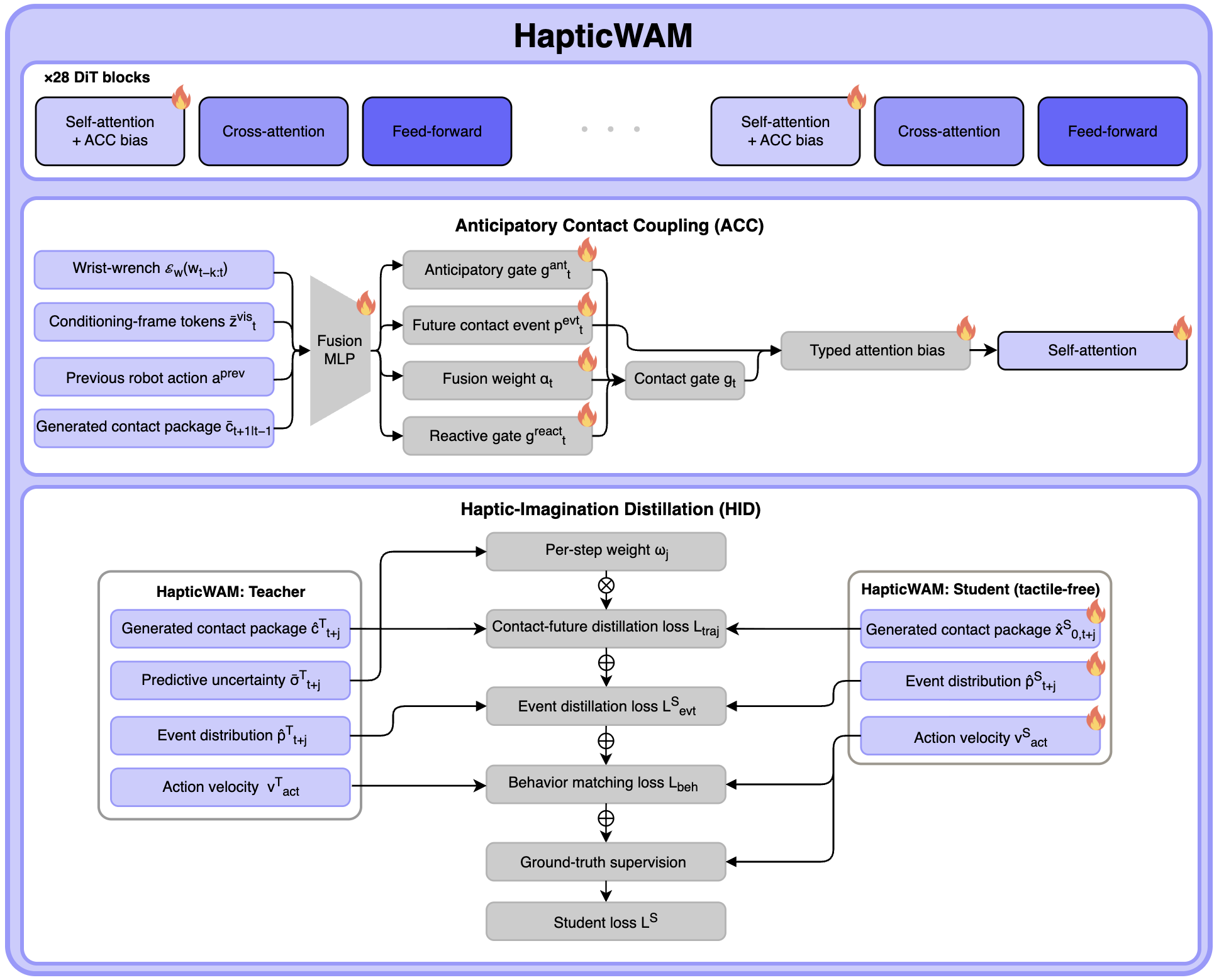}
    \caption{HapticWAM internals. \emph{Backbone:} the 28 DiT blocks of the frozen video model, each with self-attention, cross-attention to the task-text embedding and a feed-forward MLP. \emph{ACC} fuses the wrench encoding, a pooled visual summary, the previous action chunk and the contact package imagined at the previous replan into an anticipatory gate $g^{\mathrm{ant}}_t$, an event distribution $p^{\mathrm{evt}}_t$ and a blending weight $\alpha_t$; the teacher blends $g^{\mathrm{ant}}_t$ with a reactive gate from observed tactile change. The student uses the anticipatory branch alone. Each block adds the typed attention bias, to self-attention scores on haptic keys only. \emph{HID} samples the frozen teacher for contact packages, event distributions, uncertainty and action velocities, and trains the student by contact-future matching weighted by event saliency and teacher uncertainty, event distillation and action-velocity matching.}
    \label{fig:arch_detailed}
\end{figure*}

\subsection{Structured Contact Prediction}
\label{sec:method-ace}
For each of the $J=3$ future latent steps $j=1,\dots,J$, the contact
package contains a typed event $e$; per-finger changes in the displacement
field $\Delta D^f$ and normal-force field $\Delta F^{z,f}$ (superscript
$z$: the normal component); a contact mask $M^f$, centre of pressure
$\xi^f$, slip score $s^f$ and resultant wrench $F^f$; and the arm-side
wrench $w$ (time indices are omitted inside the tuple):
\begin{equation}
c_{t+j}=(e_{t+j},\{\Delta D^f,\Delta F^{z,f},M^f,\xi^f,s^f,F^f\}_f,w_{t+j}).
\label{eq:cpk}
\end{equation}
Events $e\in\{\texttt{none},\texttt{onset},\texttt{hold},\texttt{slip},\texttt{release}\}$
distinguish no contact, onset, hold, slip and release. Targets are
derived from thresholded recorded fields on the latent-time grid, without
manual contact annotation. Predicting changes emphasizes contact transients
rather than reconstructing unchanged tactile appearance.

A deterministic codec packs fields and masks on the latent grid, events
as one-hot bands, centres of pressure as Gaussian bumps, slip as tiled
values and wrenches as spatial blocks. Undefined centres of pressure
produce no bump. Actions are packed as four time steps per frame.
Event and uncertainty heads $\psi_e$ and $\psi_\sigma$ read the contact-frame
hidden states $h^c$:
\begin{align}
\hat p&=\operatorname{softmax}(\psi_e(h^c)),\label{eq:heads}\\
\log\sigma&=\psi_\sigma(h^c).\label{eq:uncertainty-head}
\end{align}
Events and mechanical channels are generated jointly; the event head is
not a separate sampling stage. A structural mask blocks direct attention
from contact/action queries to generated-video keys. Shared weights and
unmasked observation rows still permit indirect coupling.

Contact reconstruction uses a heteroscedastic loss over the output groups
$m$ (displacement, normal force, mask, centre of pressure, slip, fingertip
wrench and arm-side wrench), each with its own predicted uncertainty
$\sigma_m$:
\begin{align}
d_m&=\|\hat x_0^{(m)}-x_0^{(m)}\|_2^2,\label{eq:contact-residual}\\
\mathcal L_{\mathrm{con}}
&=\sum_m(d_m\sigma_m^{-2}+\log\sigma_m^2)
       \operatorname{sg}[\sigma_m^2]^\nu .\label{eq:contact-loss}
\end{align}
Here $\operatorname{sg}$ stops gradients and the exponent $\nu\in[0,1]$
moderates inverse uncertainty weighting. The teacher objective, with scalar
loss weights $\lambda_{(\cdot)}$ given in
\eqref{eq:train-weights}--\eqref{eq:aux-weights}, is
\begin{equation}
\begin{aligned}
\mathcal L^{\mathrm T}={}&
\lambda_a\mathcal L_{\mathrm{act}}+\lambda_v\mathcal L_{\mathrm{vid}}
+\lambda_c\mathcal L_{\mathrm{con}}+\lambda_w\mathcal L_w\\
&+\lambda_e(\mathcal L_{\mathrm{evt}}+\mathcal L_{\mathrm{band}})
+\mathcal L_{\mathrm{ACC}}
+\lambda_\sigma\|\log\sigma\|_2^2 .
\end{aligned}
\label{eq:teacher-loss}
\end{equation}
Action and video terms match flow velocities; event terms supervise the
readout and packed event band; the wrench term reconstructs the arm-side
signal. Failed demonstrations have zero ground-truth action weight.

\subsection{Anticipatory Contact Coupling}
\label{sec:method-acc}
ACC combines current leading signals with the contact package imagined at
the previous replan. A fusion MLP $\phi$ produces the representation
\begin{equation}
\mathbf u_t=\phi\!\left(
\left[\begin{array}{c}
\mathcal E_w(w_{t-k:t})\\
\bar z_t^{\mathrm{vis}}\\
\operatorname{MLP}(a^{\mathrm{prev}})\\
\operatorname{MLP}(\bar c_{t+1\mid t-1})
\end{array}\right]\right).
\label{eq:acc-e}
\end{equation}
The visual summary pools conditioning tokens; the teacher additionally
uses tactile tokens. The contact summary is aligned to the next replan
from the previously generated package, rather than taken from the ground
truth. Learned projections $W_g$ and $W_e$ give the anticipatory gate and
event distribution
\begin{align}
g_t^{\mathrm{ant}}&=\operatorname{sigmoid}(W_g\mathbf u_t),\label{eq:acc-heads}\\
p_t^{\mathrm{evt}}&=\operatorname{softmax}(W_e\mathbf u_t).\label{eq:acc-events}
\end{align}
The teacher blends this gate with a reactive gate derived from observed
tactile change:
\begin{equation}
g_t=\alpha_tg_t^{\mathrm{ant}}+(1-\alpha_t)g_t^{\mathrm{react}}.
\label{eq:acc-gate}
\end{equation}
The student uses only the anticipatory branch. In each DiT block
$\ell$, the pre-softmax attention score $r^{(\ell)}_{i,n}$ between query
$i$ and haptic key $n$ receives a typed bias:
\begin{align}
r_{i,n}^{(\ell)}&\leftarrow r_{i,n}^{(\ell)}
  +\mu_\ell\, b(p_t^{\mathrm{evt}})\,g_t,\label{eq:acc-bias}\\
b(p)&=\sum_e p_e\operatorname{softplus}(\zeta_e),\label{eq:acc-event-weight}
\end{align}
with learned per-event parameters $\zeta_e$ and per-block scales $\mu_\ell$
that start at zero. Haptic keys comprise tactile/contact frames
for the teacher and contact/proprioceptive frames for the student.
Contact and event labels supervise ACC through binary cross-entropy (BCE)
and cross-entropy (CE):
\begin{equation}
\mathcal L_{\mathrm{ACC}}=
\lambda_g\operatorname{BCE}(g_t,y_t)
+\lambda_e\operatorname{CE}(p_t^{\mathrm{evt}},e_{t+1}),
\label{eq:acc-loss}
\end{equation}
where $y_t\in\{0,1\}$ indicates contact in the labeled future interval.
The teacher's reactive gate $g_t^{\mathrm{react}}$ is a learned sigmoid MLP
of mean absolute tactile-field change; its blending coefficient $\alpha_t$
is a sigmoid projection of $\mathbf u_t$.
In two-pass training, a gradient-free sample supplies the previous-package
input for both policies; the inner pass uses a noisy target summary to
terminate recursion. Deployment reuses the previous prediction, or a
self-generated package at the first replan, without future measurements.
At deployment, ACC changes attention internally; a separate scripted
contact-probability veto controls gripper closure. No physical anticipation
lead time is inferred from this architectural design. Figure~\ref{fig:arch_detailed} summarizes the architecture of ACC and HID introduced further in Sec.~\ref{sec:method-hid}.

\begin{algorithm}[t]
\caption{Haptic-Imagination Distillation}
\label{alg:hid}
\begin{algorithmic}[1]
\Require Frozen tactile teacher; demonstrations and mixed rollouts
\State Initialize the student from teacher weights
\State Remove student gel and mechanics observation frames
\For{each training minibatch}
  \State Construct paired teacher and student observations
  \State Sample teacher contact, events and uncertainty without gradients
  \State Draw shared action noise and a shared noise level
  \State Predict student contact and action velocities
  \State Match contact targets using \eqref{eq:hid-w}--\eqref{eq:hid-event}
  \State Match teacher action velocities using \eqref{eq:hid-beh}
  \State Mask ground-truth action loss on unsuccessful episodes
  \State Add ground-truth terms and ACC auxiliaries in \eqref{eq:student-loss}
  \State Update student adapters, encoders and heads; update weight EMA
\EndFor
\State \Return Student EMA weights for deployment
\end{algorithmic}
\end{algorithm}

\subsection{Haptic-Imagination Distillation}
\label{sec:method-hid}
The student inherits teacher weights but omits tactile observation frames.
For each training batch, the frozen teacher generates contact targets,
event distributions and predicted uncertainty. A per-step weight
$\omega_j$ combines event saliency, with coefficient $\eta$, and the
teacher uncertainty $\bar\sigma^{\mathrm T}_{t+j}$ (the mean of $\sigma_m$
over output groups) with temperature $\rho$:
\begin{equation}
\omega_j=\left(1+\eta(1-\hat p^{\mathrm T}_{t+j}[\texttt{none}])\right)
\exp(-\bar\sigma^{\mathrm T}_{t+j}/\rho).
\label{eq:hid-w}
\end{equation}
This emphasizes teacher-predicted events with lower uncertainty; it does
not assume empirically calibrated confidence. The contact loss and the
event loss, a Kullback--Leibler (KL) divergence, are
\begin{equation}
\mathcal L_{\mathrm{traj}}=\sum_j \omega_j
\|\hat x^{\mathrm S}_{0,t+j}-\operatorname{pack}(\hat c^{\mathrm T}_{t+j})\|_2^2,
\label{eq:hid-traj}
\end{equation}
\begin{equation}
\mathcal L_{\mathrm{evt}}^{\mathrm S}
=\mathbb E[\operatorname{KL}(\hat p^{\mathrm T}\Vert\hat p^{\mathrm S})].
\label{eq:hid-event}
\end{equation}
Only trajectory matching uses the event/uncertainty weights. Action
distillation compares velocities at the same noisy chunk and noise level:
\begin{equation}
\mathcal L_{\mathrm{beh}}=
\mathbb E\|v_{\mathrm{act}}^{\mathrm S}(x_\tau,\tau,o^{\mathrm S})
-v_{\mathrm{act}}^{\mathrm T}(x_\tau,\tau,o^{\mathrm T})\|_2^2 .
\label{eq:hid-beh}
\end{equation}
The complete student objective retains reduced-weight ground-truth
supervision and the ACC auxiliaries:
\begin{equation}
\begin{aligned}
\mathcal L^{\mathrm S}={}&\mathcal L_{\mathrm{traj}}
+\mathcal L_{\mathrm{evt}}^{\mathrm S}
+\mathcal L_{\mathrm{beh}}+\mathcal L_{\mathrm{ACC}}\\
&+\lambda_{\mathrm{gt}}
(\mathcal L_{\mathrm{act}}+\mathcal L_{\mathrm{vid}}+\mathcal L_w)
+\lambda_\sigma'\mathcal L_{\mathrm{con}}^{\mathrm{ro}} .
\end{aligned}
\label{eq:student-loss}
\end{equation}
The final term (superscript ro: readout) trains the uncertainty readout on
detached hidden states.
Training mixes demonstrations with recorded teacher and earlier-student
rollouts; teacher matching applies even where unsuccessful rollouts have
zero ground-truth action weight. We deploy an exponential moving average
(EMA) of the student weights, which smooths the parameter trajectory across
updates and stabilizes the distilled policy. At inference, the student predicts its
own contact package without the teacher or fingertip observations.

Algorithm~\ref{alg:hid} summarizes training with the teacher and foundation
backbone frozen. The deployment intervention in Sec.~\ref{sec:ablation}
changes generated contact, not the training objective.

%% file: Experiments.tex
\subsection{Experimental Setup}
\label{sec:exp-setup}
\begin{figure}[t]
\centering
\includegraphics[width=\columnwidth]{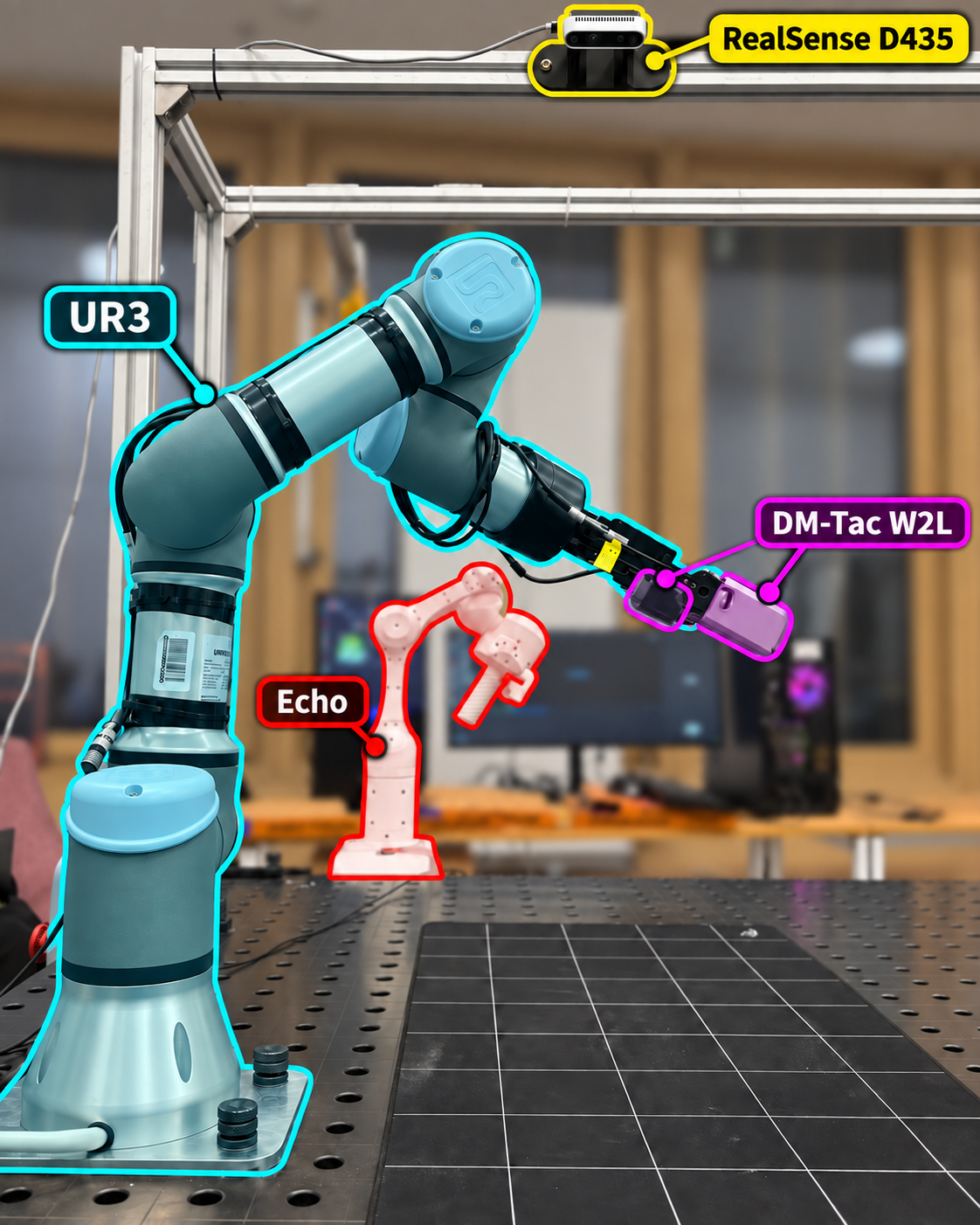}
\caption{Our setup consists of a UR3 manipulator with two DM-Tac W2L optical-tactile sensors mounted on the fingers of a Robotiq 2F-85 parallel gripper. We use Realsense D435 to collect scene RGB and Echo to teleoperate the UR3.}
\label{fig:setup}
\end{figure}

We use a UR3 with a Robotiq 2F-85 parallel gripper, two DM-Tac W2L optical
fingertips and a static RealSense D435 (Fig.~\ref{fig:setup}). We use Echo teleoperation setup to collect dataset~\cite{bazhenov2025echo}. Each fingertip
provides gel images and mechanical fields on a 384-by-288 grid over a
36-by-27\,mm surface. Tactile recording is capped at 8\,Hz, synchronized robot
commands run at 125\,Hz and we collect RGB frames from Realsense at 15\,Hz.

\subsection{Tasks and Dataset}
\label{sec:dataset}
\label{sec:exp-data}
We evaluate three tasks (Fig.~\ref{fig:teaser}):
\begin{itemize}
\item \textbf{Waffles:} pick the waffles and place them into the box.
\item \textbf{Carton:} pick the carton and place it into the box.
\item \textbf{Egg:} pick the egg and place it on the tray.
\end{itemize}
Each task contributes 250 nominal demonstrations and 35 deliberate-failure
demonstrations in which the operator intentionally over- or under-grasps the
object, driving the grasp past its feasible force range so that the episode
ends in crushing. This gives 285 episodes per task and 855 in
total, split into 761 training and 94 validation episodes. Validation
episodes never enter training and are drawn only from the nominal set, so no
deliberate failure is ever scored. We keep the failure demonstrations because
they populate regions of the contact distribution that successful
demonstrations never visit; they are supervised by the contact and video
terms with the ground-truth action loss disabled, so they shape what the
model predicts about contact without teaching it to reproduce the failing
action.
Task text accompanies all robot comparisons.

\begin{figure}[t]
\centering
\includegraphics[width=\columnwidth]{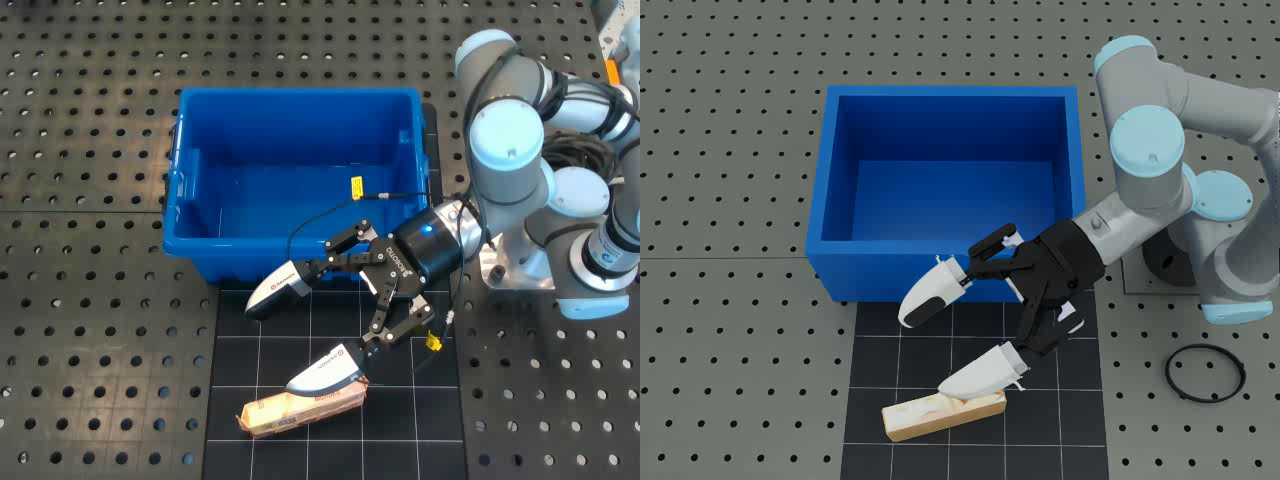}
\caption{\textbf{(Left)} Real-world setup with UR3, Robotiq gripper and DM-Tac W2L sensors. \textbf{(Right)} Simulation setup we created in NVIDIA Isaac Sim.}
\label{fig:sim}
\end{figure}

\subsection{Training and Inference}
\label{sec:exp-training}
The Cosmos-Predict2.5-2B backbone, VAE and text encoder are frozen.
Rank-16 LoRA adapters and new modules give 27.9\,M trainable teacher
parameters and 26.4\,M student parameters. Each training clip contains the
current RGB frame and 12 future targets at 4\,Hz, spanning 3\,s; these are
not 13 observed history frames. Temporal compression gives one conditioning
video latent and three future latents. Contact targets lie 1, 2 and 3\,s
after the anchor, while the 16 action targets span 0.1--1.6\,s at 10\,Hz.
At inference only the current image is observed; future video is generated.

\textbf{Teacher training.}
We use AdamW~\cite{loshchilov2017decoupled} in bfloat16 for 20{,}000 updates on one 24\,GB GPU, with effective
batch size 8, learning rates of $10^{-4}$ for LoRA and $3\cdot10^{-4}$ for
new modules, 500 warm-up updates, cosine decay and EMA decay 0.999.
Contact reconstruction uses a weighting exponent of 0.5.
The objective weights, including the ACC gate weight, are
\begin{align}
(\lambda_a,\lambda_v,\lambda_c)&=(1,0.1,1),\label{eq:train-weights}\\
(\lambda_e,\lambda_w,\lambda_g,\lambda_\sigma)&=(0.5,0.5,0.2,0.01).\label{eq:aux-weights}
\end{align}

\textbf{Simulation fine-tuning.}
The deployed teacher receives 2{,}000 additional updates on 99 scripted
simulation episodes and 203 robot rollouts, yielding 2{,}088 windows.
The simulation set includes 98 completed placements and one episode ending
with the object still gripped inside the box. We use batch size 8, learning
rates reduced fivefold and EMA decay 0.995. Simulation provides contact
event labels but no optical tactile measurements; unavailable contact
reconstruction, event-band and wrist-wrench losses are masked.
Figure~\ref{fig:sim} shows the corresponding real and simulated scenes.

\textbf{Student training.}
The deployed student is initialized from the fine-tuned teacher and distilled
for 1{,}000 updates on one H100, with effective batch size 8.
Its dataset contains 860 episodes: the 761 training demonstrations plus
99 earlier rollouts, comprising 66 teacher and 33 student episodes.
Eight windows per episode yield 6{,}880 training windows. Frozen-teacher
targets are sampled per batch; failed rollouts receive no ground-truth
action weight. Distillation uses
\begin{equation}
(\eta,\rho,\lambda_{\mathrm{gt}},\lambda_\sigma')=(4,1,0.5,1).
\label{eq:student-weights}
\end{equation}
Table~\ref{tab:genealogy} summarizes the deployed Teacher-Student pair.

\begin{table}[t]
\caption{Deployed model training.}
\label{tab:genealogy}
\centering
\small
\setlength{\tabcolsep}{3pt}
\begin{tabular}{@{}lp{5.6cm}@{}}
\toprule
Model & Initialized from, then trained \\
\midrule
HapticWAM: Teacher & Cosmos-Predict2.5-2B; 20{,}000 demonstration updates, then 2{,}000 simulation and rollout updates \\
\addlinespace[2pt]
HapticWAM: Student & Deployed teacher; 1{,}000 HID updates on demonstrations and mixed rollouts \\
\bottomrule
\end{tabular}
\end{table}

\textbf{Baselines.}
We compare against $\pi_{0.5}$~\cite{intelligence2025pi05visionlanguageactionmodelopenworld}
and Diffusion Policy~\cite{chi2023diffusion}, trained on the same 876-episode
baseline export, excluding deliberate failures. They receive neither simulation
episodes nor robot rollouts. Training uses 20{,}000 updates at batch size 6
for $\pi_{0.5}$ and 100{,}000 at batch size 64 for Diffusion Policy (DP).
The former updates 430.1\,M parameters within a 3.62\,B-parameter model;
the latter trains all 263.0\,M parameters from scratch. DP's
deployment history is 0.486\,s apart rather than the 0.1\,s training interval.

\textbf{Inference.}
On an RTX~5090 32\,Gb, HapticWAM samples four candidate chunks, screens them using
predicted contact/descent and selects by continuity with the previous plan.
Median replan times over the 200 trials are 0.445\,s for the teacher (compiled kernels), 0.636\,s for the student (uncompiled), 0.127\,s for $\pi_{0.5}$ and 0.389\,s for DP. HapticWAM additionally uses a contact-based closing veto and
uncertainty-dependent playback; baseline playback is unscaled.

\subsection{Evaluation Protocol}
\label{sec:exp-closedloop}
Each of four models is evaluated on 20 waffle, 20 carton and 10 egg starts,
for 200 robot trials. The shared executor retains the same safety guards,
but candidate selection, gripper timing and replan limits differ across models.
UR3 starting positions are shared across models using seeds 101--120
for waffles/carton and 101--110 for eggs, with model order alternated
between launches. Success is an operator-reviewed placement of the object at the target. Independently, every trial receives a sensor-based grasp-quality label from the recorded fingertip force: an over-grasp when the pinch peak exceeds the force reference of Eq.~\eqref{eq:force-reference}, an under-grasp when the fingers contact the object but no held lift follows, and clean otherwise; on eggs, damage is scored by shell cracking. The sensor rule agrees with the operator on placement in 154 of the 160 waffle and carton trials; the operator was not blinded to the running policy. We report the binary
placement rate:
\begin{equation}
\mathrm{Success\ rate}=\frac{100}{N}\sum_{i=1}^{N}
\mathbf 1\{\text{final verdict is placed}\}.
\label{eq:success}
\end{equation}
Force is reported separately; the placement rate is not redefined by a
post-hoc force threshold. Because no independently calibrated damage
threshold exists for the waffle and carton, we derive a descriptive force
reference from the evaluation trials themselves. Let $\mathcal P$ be the set
of per-trial pinch peaks, the maximum resultant fingertip force recorded
during a trial, over the 73 waffle and carton trials that the sensor rule
scores as placements, pooled across all four models. With $\mu_{\mathcal P}$
and $s_{\mathcal P}$ its sample mean and sample standard deviation, the
reference is the three-sigma upper envelope of successful-placement forces,
\begin{equation}
F_{\mathrm{ref}}=\mu_{\mathcal P}+3s_{\mathcal P}=24.8\,\mathrm N.
\label{eq:force-reference}
\end{equation}
A pinch peak above $F_{\mathrm{ref}}$ is therefore a force atypical of a
successful placement of these objects; we call it a reference exceedance
rather than damage, since neither object visibly breaks. Eggs are handled
differently for two reasons: their damage is directly observable as shell
cracking, so no force proxy is needed, and their tolerable force range
differs from that of the waffle and carton, so their trials do not enter
$\mathcal P$ and their over-grasp label is not defined by $F_{\mathrm{ref}}$;
egg forces are still reported, but only for description. Reported forces use
the resultant-force readout in newtons under the configured sensor scale,
not the uncalibrated per-pixel force field.

\subsection{Main Results}
\label{sec:results-main}
Table~\ref{tab:rigpairs} compares the deployed teacher, its distilled student,
$\pi_{0.5}$ and DP. The student achieves the highest observed
placement rate on each task: 95\% on waffles, 85\% on cartons and 50\% on eggs. Paired exact sign tests over shared starts (four-level outcome: no held grasp, held, lifted, placed) put the student above $\pi_{0.5}$ on every task ($p\le0.016$) and above DP on waffles and cartons ($p<0.001$; eggs $p=0.69$); the student--teacher gap is not separable at this size ($p=0.062$, $0.125$ and $1.0$).
Across the 50 starts, it places 41 objects, versus 30 for the teacher,
10 for $\pi_{0.5}$ and 9 for DP.
The largest student--teacher differences occur on waffles and cartons.
Egg handling remains challenging for all methods, and its smaller trial
count limits precision: the student's 50\% egg success has a descriptive
95\% Wilson interval of 23.7--76.3\%. Task-wise 95\% Wilson intervals are 76.4--99.1\% (waffles), 64.0--94.8\% (cartons) and 23.7--76.3\% (eggs). These observations concern the complete
configurations described above; they do not isolate a contact-distillation
effect from training-data and execution differences.

\begin{table}[t]
\caption{Pick-and-place success rate: operator placement verdicts.}
\label{tab:rigpairs}
\centering
\small
\setlength{\tabcolsep}{4pt}
\begin{tabular}{lccc}
\toprule
Model & Waffles $\uparrow$ & Carton $\uparrow$ & Egg $\uparrow$ \\
\midrule
\textbf{Teacher (Ours)} & 70\% (14/20) & 60\% (12/20) & 40\% (4/10) \\
\rowcolor{oursrow}
\textbf{Student (Ours)} & \textbf{95\% (19/20)} & \textbf{85\% (17/20)} & \textbf{50\% (5/10)} \\
\addlinespace[2pt]
$\pi_{0.5}$ & 20\% (4/20) & 30\% (6/20) & 0\% (0/10) \\
Diffusion Policy & 10\% (2/20) & 15\% (3/20) & 40\% (4/10) \\
\bottomrule
\end{tabular}
\end{table}

\begin{table*}[t]
\caption{Pinch force at sensor-defined placements, mean and standard deviation.}
\label{tab:force}
\centering
\small
\setlength{\tabcolsep}{8pt}
\begin{tabular}{lrrrrrr}
\toprule
& \multicolumn{2}{c}{Waffles} & \multicolumn{2}{c}{Carton} & \multicolumn{2}{c}{Egg} \\
\cmidrule(lr){2-3} \cmidrule(lr){4-5} \cmidrule(lr){6-7}
Model & Force (N) & Samples & Force (N) & Samples & Force (N) & Samples \\
\midrule
\textbf{Teacher (Ours)} & $16.7\pm4.7$ & 13 & $11.4\pm1.4$ & 12 & $24.9\pm9.2$ & 5 \\
\textbf{Student (Ours)} & $14.3\pm3.6$ & 19 & $11.8\pm2.2$ & 16 & $18.8\pm9.1$ & 6 \\
\addlinespace[2pt]
$\pi_{0.5}$ & $13.4\pm2.0$ & 5 & $16.0\pm4.9$ & 5 & --- & 0 \\
Diffusion Policy & $14.1$ & 1 & $10.1$ & 2 & $27.1\pm3.1$ & 4 \\
\bottomrule
\end{tabular}
\end{table*}

\subsection{Contact-Imagination Ablation}
\label{sec:ablation}
\textbf{Intervention.} We evaluate the same deployed student checkpoint
without further training, replacing the generated contact frames with a
zero contact package throughout sampling. The checkpoint weights and
available observation channels are unchanged. The intervention clamps the contact
representation, rather than removing a physical sensor or retraining the
student. We compare 30 additional robot trials against the intact student's
previously recorded outcomes on the same first ten starts per task.
Both conditions use the final operator placement verdict.

\begin{table}[t]
\caption{Contact-imagination ablation: placement success rate.}
\label{tab:imagination}
\centering
\small
\setlength{\tabcolsep}{5pt}
\begin{tabular}{lcc}
\toprule
Task & Intact student $\uparrow$ & Zero contact $\uparrow$ \\
\midrule
Waffles & 90\% (9/10) & 10\% (1/10) \\
Carton & 70\% (7/10) & 30\% (3/10) \\
Egg & 50\% (5/10) & 0\% (0/10) \\
\midrule
All tasks & 70\% (21/30) & 13.3\% (4/30) \\
\bottomrule
\end{tabular}
\end{table}

\textbf{Results.} Clamping contact frames reduces placements on every task
(Table~\ref{tab:imagination}), from 21 to 4 over the 30 shared starts.
The absolute success-rate reduction is 56.7 percentage points, with the
largest task-level drop on waffles. Both columns are recorded trials scored by the same operator placement verdict; the intact column is the first ten starts per task of Table~\ref{tab:rigpairs}.

\textbf{Interpretation.} The result supports a deployment-time dependence
on the contact-generation pathway in the evaluated system, and an offline
probe on recorded demonstrations points the same way. On the 94 held-out
validation episodes we sample the student as at deployment during the last
1.5\,s before the demonstrator first closes the gripper, and measure the
\emph{terminal endpoint error}: the distance between the hand position the
predicted 1.6\,s action chunk would reach and the position the demonstrator
actually reached. A policy that does not move scores 28.2\,mm. The intact
student scores 22.31\,mm; with its contact frames clamped to zero it scores
28.31\,mm, no better than not moving, whereas zeroing only the ACC input
leaves it at 22.35\,mm. The harm therefore comes from the content of the
generated contact frames rather than from the ACC input, so the comparison
is not an isolated test of ACC's attention bias, nor a measure of
contact-prediction accuracy. It also does not isolate the mechanism on the
robot: clamping changes the jointly generated representation, and
contact-dependent selection and closing rules can also mediate the executed
outcome. Nor does it establish that contact supervision is better than
action-only distillation: both conditions use the same HID-trained weights.
The intact trials precede the intervention trials, so matching start seeds
does not remove session-order effects.

\subsection{Force Analysis}
\label{sec:force-results}
Table~\ref{tab:force} reports pinch force conditional on sensor-defined
placement; these samples can differ from the final operator-reviewed
placement counts. The student has lower mean force than the teacher on
waffles and eggs, with similar carton forces. Four reference exceedances
occurred across the waffle/carton trials: two for the teacher on waffles
and two for $\pi_{0.5}$ on cartons. None was recorded for the student. Under-grasps (contact without a held lift): teacher 5/1/2, student 0/2/1, $\pi_{0.5}$ 7/3/4 and DP 10/1/1 on waffles/cartons/eggs; the student alone has neither an over- nor an under-grasp on waffles. The reference is an in-sample waffle/carton band, not calibrated for eggs, where every model exceeds it.
Two egg cracks were recorded for DP and none for the other
models. These descriptive observations do not establish damage-free control:
force summaries exclude failed placements and event detection is limited
by sensor coverage.

Force entries show mean and standard deviation where reported; the
single-sample and two-sample DP summaries retain only the mean.

\subsection{Student--Teacher Gap}
\label{sec:gap}
The tactile-free student places more objects than the tactile teacher it
was distilled from (41 versus 30 of 50). Three properties of HID make this direction
plausible rather than paradoxical.

\textbf{Corrective on-policy targets.} The student's 99 rollout episodes
include 33 of its own. On failed rollouts the ground-truth action loss is
masked, but Eq.~\eqref{eq:hid-beh} still supplies the frozen teacher's
velocity $v^{\mathrm T}_{\mathrm{act}}$ at every visited state. This is a
DAgger-style expert query~\cite{ross2011dagger} with the teacher as expert:
the student is corrected on the states it actually reaches at deployment,
whereas no comparable expert exists for the teacher, whose rollouts can only
be relabeled by outcome. The reversal between offline endpoint error,
measured on demonstration states, and closed-loop placement is the expected
signature of such a covariate-shift correction.

\textbf{A tactile-independent contact gate.} The student fixes
$\alpha_t\equiv1$ in Eq.~\eqref{eq:acc-gate}, so its attention bias depends
only on the anticipatory summary,
\begin{equation}
g^{\mathrm S}_t=g^{\mathrm{ant}}_t=\operatorname{sigmoid}(W_g\mathbf u_t),
\label{eq:gap-gate}
\end{equation}
whereas the teacher's gate also tracks $g^{\mathrm{react}}_t$, a function of
tactile-field change sampled at 8\,Hz (Sec.~\ref{sec:exp-setup}) that
includes the uncalibrated per-pixel force field. The student cannot be
perturbed by this channel. Consistent with this, on waffles the teacher
records five under-grasps and two reference exceedances where the student
records none (Sec.~\ref{sec:force-results}).

\textbf{Uncertainty-filtered targets.} With $(\eta,\rho)=(4,1)$ from
Eq.~\eqref{eq:student-weights}, the per-step weight of Eq.~\eqref{eq:hid-w}
lies in
\begin{equation}
\omega_j\in\left[e^{-\bar\sigma^{\mathrm T}_{t+j}},\,
5e^{-\bar\sigma^{\mathrm T}_{t+j}}\right],
\label{eq:gap-weight}
\end{equation}
so a teacher contact prediction at unit uncertainty carries at most $0.37$
times the weight of a confident one, and predicted events up to five times
the weight of quiescent steps. The student therefore regresses a
selectively confident, event-emphasized version of the teacher rather than
the teacher itself~\cite{furlanello2018born}; with
$\lambda_{\mathrm{gt}}=0.5$ ground-truth anchoring, this can smooth the
teacher's errors on precisely the steps where it is unsure.

%% file: Conclusion.tex
HapticWAM combines heterogeneous tactile encoding, structured contact
generation and contact-conditioned attention within a world--action model.
Haptic-Imagination Distillation transfers contact and action predictions
to a student without fingertip tactile inputs. Across the evaluated tasks,
the student achieves a 77\% per-task mean pick-and-place success rate (82\% pooled) versus 57\% (60\% pooled) for the teacher, and the highest observed placement rate among the evaluated configurations. The common execution controller retains tactile safeguards.
Clamping its generated contact frames to zero reduces placement success
on shared starts, supporting deployment-time dependence on this pathway.
The incremental training benefit of contact supervision over action-only
distillation remains to be isolated.